\documentclass[letterpaper]{article} 
\usepackage{aaai2026}  
\usepackage{times}  
\usepackage{helvet}  
\usepackage{courier}  
\usepackage[hyphens]{url}  
\usepackage{graphicx} 
\usepackage{natbib}  
\usepackage{caption} 
\usepackage{algorithm}
\usepackage{algorithmic}

\usepackage{newfloat}
\usepackage{listings}
\DeclareCaptionStyle{ruled}{labelfont=normalfont,labelsep=colon,strut=off} 
\floatstyle{ruled}
\newfloat{listing}{tb}{lst}{}
\floatname{listing}{Listing}
\usepackage{amsmath}
\usepackage{booktabs}
\usepackage{bigstrut}
\usepackage{rotating}
\usepackage{multicol}
\usepackage{multirow}
\usepackage{overpic}
\usepackage{subfigure}
\usepackage{amssymb}
\usepackage{comment}
\usepackage{enumitem}

\title{Scale-Disentangled spatiotemporal Modeling for Long-term Traffic Emission Forecasting}
\author{
    Yan Wu\textsuperscript{\rm 1},
    Lihong Pei\textsuperscript{\rm 2},
    Yukai Han\textsuperscript{\rm 1},
    Yang Cao\textsuperscript{\rm 1},
    Yu Kang\textsuperscript{\rm 1},
    YanlongZhao\textsuperscript{\rm 2}\\
}
\affiliations{
     \textsuperscript{\rm 1}Department of Automation, University of Science and Technology of China\\
     \textsuperscript{\rm 2}State Key Laboratory
of Mathematical Sciences, Academy of Mathematics and Systems Science, Chinese Academy of Sciences\\

    wuyanobsidian@mail.ustc.edu.cn,
    plh1225@amss.ac.cn,
    hanyukai@mail.ustc.edu.cn, \\
    forrest@ustc.edu.cn,
    kangduyu@ustc.edu.cn,
    ylzhao@amss.ac.cn
}

\usepackage{bibentry}

\begin{document}

\maketitle

\begin{abstract}
Long-term traffic emission forecasting is crucial for the comprehensive management of urban air pollution. Traditional forecasting methods typically construct spatiotemporal graph models by mining spatiotemporal dependencies to predict emissions. However, due to the multi-scale entanglement of traffic emissions across time and space, these spatiotemporal graph modeling method tend to suffer from cascading error amplification during long-term inference. To address this issue, we propose a \underline{S}cale-\underline{D}isentangled \underline{S}patio-\underline{T}emporal \underline{M}odeling (SDSTM) framework for long-term traffic emission forecasting. It leverages the predictability differences across multiple scales to decompose and fuse features at different scales, while constraining them to remain independent yet complementary. Specifically, the model first introduces a dual-stream feature decomposition strategy based on the Koopman lifting operator. It lifts the scale-coupled spatiotemporal dynamical system into an infinite-dimensional linear space via Koopman operator, and delineates the predictability boundary using gated wavelet decomposition. Then a novel fusion mechanism is constructed, incorporating a dual-stream independence constraint based on cross-term loss to dynamically refine the dual-stream prediction results, suppress mutual interference, and enhance the accuracy of long-term traffic emission prediction. Extensive experiments conducted on a road-level traffic emission dataset within Xi’an’s Second Ring Road demonstrate that the proposed model achieves state-of-the-art performance.

\end{abstract}


\section{Introduction}
The rapid pace of urbanization and the continuous aggravated in the number of motor vehicles have intensified the problem of urban traffic pollution. Vehicular pollutants, such as carbon monoxide (CO) and nitrogen oxides (NOx), bring significant threats to the ecological environment and public health. Developing scientific and accurate models for vehicle emission forecasting are essential for effectively capturing the dynamic patterns of pollution within traffic systems, for they can provide critical data guidance and decision support for traffic flow regulation and air pollution control.

With the rapid development of data mining technologies, striking progress has been made in spatiotemporal prediction.
Spatiotemporal graph neural networks(STGNNs) have been widely used to capture the relationships and their evolution in non-Euclidean space.
STGCN \cite{yu2017spatio} applies Graph Convolutional Network to spatiotemporal forecasting.
GWNet \cite{wu2019graph} and AGCRN \cite{bai2020adaptive} captures hidden spatial correlations through adaptive adjacency matrixs.
STGODE \cite{fang2021spatial} uses Graph Ordinary Differential Equations (ODEs) to model the dynamic interactions between nodes.
STFNN \cite{feng2024spatio} models air quality as a spatiotemporal field and handles continuous data by learning the gradient of the field.
FasterSTS \cite{dai2025fastersts} combines trajectory graph modeling with structure-aware Transformer to improve the accuracy of multi-agent trajectory prediction.
LightST \cite{zhang2025efficient} accelerates traffic forecasting and alleviates the over-smoothing in GNNs through knowledge distillation.
ST-FiT \cite{lei2025st} performs data augmentation separately in time and space and combines it with an iterative strategy to deal with limited training data.
Transformer-based method has earned success in forecasting.
NRFormer \cite{lyu2024nrformer} captures complex nuclear radiation spatio-temporal dynamics by integrating non-stationary temporal attention, imbalance-aware spatial attention, and radiation propagation prompting modules.
Graph Transfomers like GRIT \cite{ma2023graph} and STGformer \cite{wang2024stgformer} combines self-attention with structural encoding has improved expressivity.
Large language models (LLMs) have been introduced to enhance the ability of spatiotemporal models to capture long-term dependencies due to their strong contextual understanding. GATGPT \cite{chen2023gatgpt} introduces graph attention into LLMs to identify spatial dependencies.
ST-LLM \cite{liu2024spatial} enhances traffic forecasting by integrating spatiotemporal inputs with partially frozen LLMs.
UrbanMind \cite{liu2025urbanmind} introduces the fusion-masked autoencoder Muffin-MAE to capture complex spatiotemporal dependencies among urban dynamics.

Although current models have achieved notable progress in spatiotemporal forecasting, most existing models rely on the assumption of independent and identically distributed and simply fit the surface-level patterns of spatiotemporal data, neglecting the interactions and structural heterogeneity among different temporal scales (like trends and cycles) and spatial scales (like local clustering). The evolution of real-world spatiotemporal sequences exhibits multi-scale mixing characteristics, with data distribution patterns differing across hierarchies, making it hard to identify the true underlying evolutionary mechanisms. Traditional models have limited understanding of highly multi-scale coupled spatiotemporal sequences and tend to mistake high-frequency local dynamics for noise. It leads to the gradual accumulation and amplification of errors during iteration, which is particularly pronounced in long-term multi-step predictions. Consequently, it impairs the prediction accuracy and generalization ability of model, reducing its overall predictability.

In summary, due to the multi-scale entanglement of traffic emissions across time and space, modeling bias amplifies with the extension of the prediction horizon, making it hard to ensure predictability. These factors make spatiotemporal prediction a highly challenging problem, as detailed below:

\begin{itemize}
    \item The multi-scale cascading mechanism of spatiotemporal data remains ambiguous, modeling under entangled states leads to poor performance in cross-scale prediction tasks. It limits the model’s ability to identify and understand complex spatiotemporal dependencies, thereby limiting its generalization capability and robustness.
    \item The components obtained through multi-scale feature disentanglement exhibit vastly different statistical properties. The pronounced pattern differences among these components make it difficult for a single model to learn and optimize them simultaneously, and also cause difficulty in ensuring the effectiveness of their feature fusion.
\end{itemize}

To address the above challenges, it is urgent to have a modeling paradigm capable of uncovering the intrinsic patterns in the spatiotemporal evolution process. Considering the inherent multi-scale cascading characteristics of real spatiotemporal data, we propose a \underline{S}cale-\underline{D}isentangled \underline{S}patio-\underline{T}emporal \underline{M}odeling (SDSTM) framework for long-term traffic emission forecasting. We decouple the spatiotemporal dynamic evolution into multi-scale components spanning both time and space, based on the differing predictability strengths of stable and dynamic components. Specifically, we divide all possible spatiotemporal states into four quadrants: time-stable and space-stable features, time-dynamic and space-stable features, time-stable and space-dynamic features, and time-dynamic and space-dynamic features. In the original data space, however, these four states are highly coupled during the spatiotemporal evolution process. Therefore, to further enhance decomposability, we introduce the Koopman lifting operator into the decoupling framework, effectively reducing decomposition difficulty and improving the accuracy of predictability boundary construction. Additionally, considering the substantial differences in the characteristics of the decomposed components, it is crucial to ensure the independence of these components during long-term prediction to avoid weakening the strongly predictable components. Our contributions are summarized as follows:

\begin{itemize}
    \item We propose a dual-stream feature decomposition strategy integrates Koopman theory with a gated wavelet decomposition mechanism to achieve linear embedding and multi-scale dynamic disentanglement of nonlinear spatiotemporal systems, effectively demarcating the predictability boundary, thereby enhancing the model’s ability to analyze complex spatiotemporal evolution.
    \item We design a novel evidence lower bound (ELBO) fusion mechanism, introducing a cross-term loss to constrain the parallel training and collaborative representation of dual branches. This mechanism enables dynamic refinement of prediction results and decision-level fusion, thereby achieving joint optimization of the final results.
    \item We conducted comprehensive evaluation on a realistic road-level vehicle emission dataset. Results validate that SDSTM exhibits competitive performance and shows remarkable robustness to various prediction horizons.
\end{itemize}

\section{Method}
\subsection{Overview}
In this section, we provide details of the main framework of SDSTM. SDSTM adopts a hierarchical structure, consisting of multiple SDSTM blocks with identical internal architecture connected sequentially. We encourage each block to perform hierarchical learning of operators by processing the dynamic residuals fitted by the previous block, thereby progressively correcting and extracting finer dynamic patterns.

\begin{figure*}[!t]
\centering
\includegraphics[width=\textwidth]{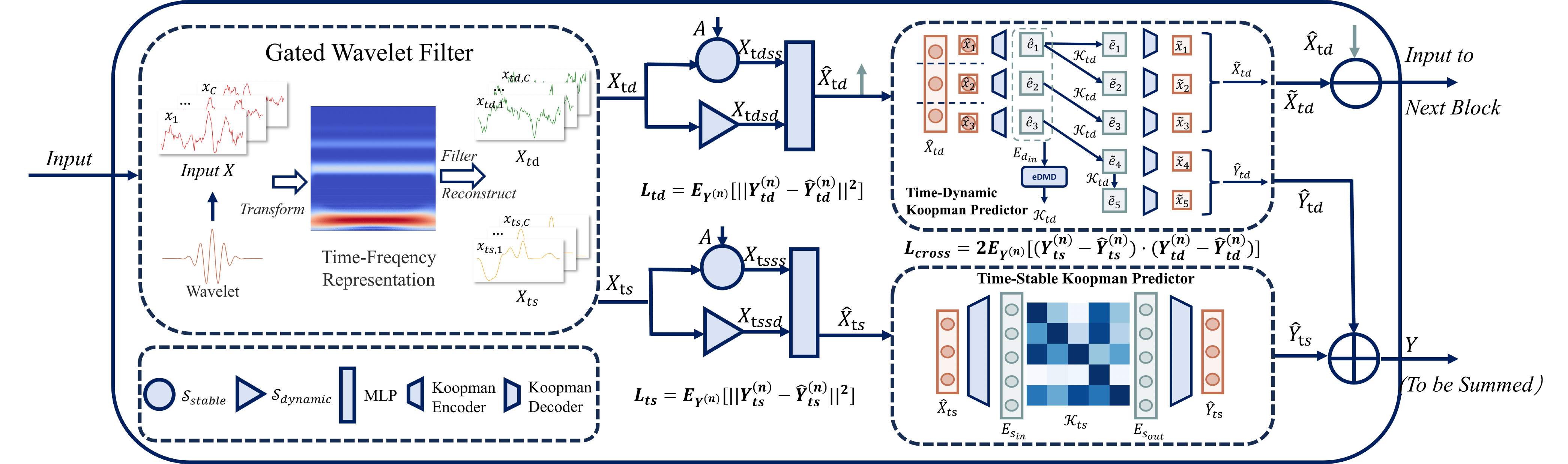}
\caption{Main structure of SDSTM blocks.}
\label{fig:method}
\end{figure*}

To address the challenges of multi-scale entanglement and structural heterogeneity in spatiotemporal forecasting, we propose a dual-stream multi-scale modeling framework based on predictability decomposition.
For the $n-$th block, the input signal is $X^{(n)}=[x_1, x_2, ......x_T]^T \in \mathbb{R}^{T\times D}$, where $T$  is the length of the historical window, i.e., the number of time steps in the input data, and $D$ is the feature dimension of $X^{(n)}$, which, in the context of exhaust emission prediction, corresponds to the number of monitoring nodes.

First, we perform frequency analysis on input $X^{(n)}$ via a gated wavelet filter. The signal is decomposed along the temporal dimension to obtain the time-stable component $X_{ts}^{(n)}$ and the time-dynamic component $X_{td}^{(n)}$ of the original input.

\begin{equation}
X^{(n)}_{ts},X^{(n)}_{td} = WaveletFilter(X^{(n)})
\end{equation}

Next, we model the explicit relationships and implicit dependencies along the spatial dimension separately, obtaining four components: \underline{t}ime-\underline{s}table \underline{s}pace-\underline{s}table components $X_{tsss}^{(n)}$, \underline{t}ime-\underline{d}ynamic \underline{s}pace-\underline{s}table components $X_{tdss}^{(n)}$, \underline{t}ime-\underline{s}table \underline{s}pace-\underline{d}ynamic components $X_{tssd}^{(n)}$, and \underline{t}ime-\underline{d}ynamic \underline{s}pace-\underline{d}ynamic components $X_{tdsd}^{(n)}$.



\begin{equation}
X_{tsss}^{(n)}= \mathcal{S}_{stable}(X_{ts}^{(n)}), X_{tdss}^{(n)} =\mathcal{S}_{stable}(X_{td}^{(n)})
\label{eq:static}
\end{equation}

\begin{equation}
X_{tssd}^{(n)}= \mathcal{S}_{dynamic}(X_{ts}^{(n)}), X_{tssd} ^{(n)}=\mathcal{S}_{dynamic}(X_{td}^{(n)})
\label{eq:dynamic}
\end{equation}

Here, $\mathcal{S}_{stable}$ and $\mathcal{S}_{dynamic}$ denote the space-stable and space-dynamic module. Then we fuse these four components to obtain the new time-stable $\hat{X}^{(n)}_{ts}$ and time-dynamic component $\hat{X}^{(n)}_{td}$, both carried with spatial information.

\begin{equation}
\begin{aligned}
\hat{X}^{(n)}_{ts}=MLP(X^{(n)}_{tsss}, X^{(n)}_{tssd}),
\hat{X}^{(n)}_{td}=MLP(X^{(n)}_{tdss}, X^{(n)}_{tdsd})
\end{aligned}
\end{equation}

We map $\hat{X}^{(n)}_{ts}$ and $\hat{X}^{(n)}_{td}$ into the Koopman embedding space, further enhancing the spatiotemporal states' predictability through feature dimension augmentation. To ensure embedding consistency, the dual-stream branches share the encoder $\phi_{enc}$ and the decoder $\phi_{dec}$. Next, we use the corresponding Koopman operators to predict them, and then map the predicted states back to the original data space. Inspired by ELBO concept, we introduce a cross-term loss to constrain the collaborative optimization of the two predictors. The time-stable predictor generates the prediction output $\hat{Y}_{ts}^{(n)}$, while the time-dynamic predictor outputs the prediction result $\hat{Y}_{td}^{(n)}$ and the reconstructed input $\tilde{X}^{(n)}_{td}$.

\begin{equation}
\hat{Y}^{(n)}_{ts} = KP_{ts}(\hat{X}^{(n)}_{ts})
\end{equation}

\begin{equation}
\tilde{X}^{(n)}_{td},\hat{Y}^{(n)}_{td}=KP_{td}(\hat{X}^{(n)}_{td})
\end{equation}

We compute the residual between the input signal and its reconstruction as the input for the next block, in order to guide and refine subsequent modeling. The final prediction result $Y$ is the sum of the prediction results from all blocks.

\begin{equation}
X^{(n+1)}=X^{(n)}_{td}-\tilde{X}^{(n)}_{td},\hat{Y}=\sum_{n}(\hat{Y}^{(n)}_{ts} +\hat{Y}^{(n)}_{td} )
\end{equation}

\subsection{Dual-Stream Feature Decomposition}
\subsubsection{Gated Wavelet Decomposition}
Realistic spatiotemporal sequences inherently exhibit non-stationarity, and the complexity that their features varying with time distribution makes the forecasting task a great challenge. A feasible treatment is to decompose the non-stationary sequences into two parts: a time-stable component with strong non-stationarity that reflects long-term trends, and a time-dynamic component capturing weak non-stationary information such as local abrupt changes or rapid fluctuations.


We choose wavelet transform for feature decoupling in time dimension. It uses a decaying wavelet basis to capture local signal features across multiple time scales via multi-resolution analysis, enabling finer temporal characterization.

We first apply the discrete wavelet transform $\mathcal{W}$ to decompose the input $X^{(n)}$ into two parts: the low-frequency coefficients $L^{(n)}$ and the high-frequency coefficients $H ^{(n)}$, where $J$ denotes decomposition levels in wavelet transform.

\begin{equation}
\mathcal{W}(X^{(n)}) = (L^{(n)},\{H ^{(n)}\}^J_{j=1})
\end{equation}

Then, we apply the inverse wavelet transform to the previously separated low-frequency part to reconstruct the signal, which serves as an initial estimate of the time-stable component to obtain the low-frequency component of the original signal, which is referred to as $X_{low}$.

\begin{equation}
X_{low}^{(n)}=\mathcal{W}^{-1}(L^{(n)},\{0\}^J_{j=1}))
\end{equation}

Here, $\mathcal{W}$ and $\mathcal{W}^{-1}$ denote the direct and the inverse discrete wavelet transform, $L^{(n)}$ and $H ^{(n)}$ represent the low-frequency and high-frequency components of $X^{(n)}$.

We compare the reconstructed low-frequency component $X_{low}^{(n)}$ with the original input signal $X^{(n)}$  to generate a gating coefficient $\gamma$, which quantifies the degree of difference between
$X_{low}^{(n)}$ and $X^{(n)}$ at each timestep, i.e., the level of non-stationarity of the signal at each timestep.

\begin{equation}
\gamma = Sigmoid(Conv1D (X^{(n)}, X_{\text{low}}^{(n)}))
\end{equation}

In fact, the low-frequency component $X_{low}^{(n)}$ slightly differs from the original input $X_{}^{(n)}$ at certain nodes, indicating that $X_{}^{(n)}$ is relatively stationary at those points, and the difference can be attributed to noise caused by measurement errors or other factors. In such cases, we reduce our attention to these points based on the gated coefficient $\gamma$ .

We utilize the gating coefficient $\gamma$ to perform a soft separation of the time-dynamic component, extracting the local detail information contained in $X^{(n)}$ which we refer to as the time-dynamic component $X_{td}^{(n)}$. By subtracting $X_{td}^{(n)}$ from $X^{(n)}$, we obtain the time-stable component $X_{ts}^{(n)}$, which describes the global trend of the original signal.

\begin{equation}
X_{td}^{(n)}=\gamma \odot (X^{(n)}-X_{low}^{(n)}), X_{ts}^{(n)}=X^{(n)}-X_{td}^{(n)}
\end{equation}

To describe the evolution patterns of traffic emissions across  spatial scales, we first learn the intuitive stable spatial dependencies among physically connected road segments based on the prior road network structure, aiming to model the long-term stable patterns within the emission system. Then, we employ a graph attention mechanism to capture the implicit dynamic spatial correlations among road segments, enhancing the model's ability to perceive local abrupt changes and non-stationary evolution processes. We feed $X_{ts}^{(n)}$ and $X_{td}^{(n)}$ respectively into the space-stable module $\mathcal{S}_{stable}$ and the space-dynamic module $\mathcal{S}_{dynamic}$ to enable targeted modeling of different spatial structural features.

\begin{equation}
X_{tsss}^{(n)}= \mathcal{S}_{stable}(X_{ts}^{(n)},\hat{A}_{\alpha}), X_{tssd} ^{(n)}=\mathcal{S}_{dynamic}(X_{ts}^{(n)})
\end{equation}

\begin{equation}
X_{tdss}^{(n)}= \mathcal{S}_{stable}(X_{td}^{(n)},\hat{A}_{\alpha}), X_{tdsd} ^{(n)}=\mathcal{S}_{dynamic}(X_{td}^{(n)})
\end{equation}

Here, $\hat{A}_{\alpha}=\tilde{D}^{-1/2}\tilde{A}\tilde{D}^{-1/2}$ denotes the symmetrically normalized adjacency matrix, $\tilde{A}_{\alpha}= \alpha A +(1-\alpha) I_N$ represents the weighted adjacency matrix with self-loops introduced, $A$ is the adjacency matrix constructed based on the prior road network structure. In this work, GCN \cite{kipf2016semi} is employed as $\mathcal{S}_{stable}$, and the attention module from DIFFormer \cite{wu2023difformer} is used as $\mathcal{S}_{dynamic}$.

To effectively fuse space-stable and space-dynamic features, we first concatenate these two input features along the channel dimension and use an MLP to generate a fusion weight $\beta \in [0,1]$, and then incorporate Squeeze-and-Excitation Networks \cite{hu2018squeeze} to adaptively adjust the importance of each component, ultimately producing a more expressive spatial fusion representation.

\begin{equation}
\begin{aligned}
    \hat{X}_{ts}^{(n)} &= \beta_{ts} \odot X_{tsss}^{(n)} + (1-\beta_{ts} )X_{tssd}^{(n)} \\
    \hat{X}_{td}^{(n)} &= \beta_{td}  \odot X_{tdss}^{(n)} + (1-\beta_{td} )X_{tdsd}^{(n)}
\end{aligned}
\end{equation}

Here, $\beta_{ts}$ and $\beta_{td}$ denote the weighting coefficient, $\hat{X}^{(n)}_{ts}$ and $\hat{X}^{(n)}_{td}$ denote the new time-stable and time-dynamic components, enriched with spatial features.

\subsubsection{Koopman Embedding}
To further enhance the accuracy of predictability boundary construction, we introduce the Koopman operator for dimension augmentation. Specifically, we use an encoder $\phi_{enc}$ to project the input data into a Koopman embedding space with higher dimension. For the time-stable component, we map it directly. For the time-dynamic component, we first divide the input $\hat{X}^{(n)}_{td}$ into $\frac{T}{L}$ subsequences $\hat{x}_i \in \mathbb{R}^{L \times C}$ with length $L$, then map them individually into the Koopman embedding space.

\begin{equation}
\begin{aligned}
E_{s_{in}}&=\phi_{enc}(\hat{X}_{ts}^{(n)}),\\
E_{d_{in}}&=[\hat{e}_1,...,\hat{e}_{\frac{T}{L}}]=[\phi_{enc}(\hat{x}_1),...,\phi_{enc}(\hat{x}_{\frac{T}{L}})]
\end{aligned}
\end{equation}

\subsection{Fusion Architecture for Prediction}
Through the dual-stream feature decomposition strategy, we separate the original data into two component based on predictability. In this section, we predict them in parallel via two types of Koopman predictor. Meanwhile, an novel ELBO fusion mechanism is designed to constrain dual branches, achieving collaborative optimization of the final results.

\subsubsection{Time-Stable Koopman Prediction}
To capture the global evolution patterns, we design a time-stable Koopman predictor $KP_{ts}$, which learns the Koopman operator $\mathcal{K}_{ts}\in \mathbb{R}^{D \times D}$ to directly describe the transition between historical and predicted sequences, where $D$ denotes the dimension of Koopman embedding space. Then we use the decoder $\phi_{dec}$ to decode the predicted states in Koopman space back to the original space, obtaining the predicts of the time-stable component.

\begin{equation}
E_{s_{out}}=\mathcal{K}_{ts}E_{s_{in}}, \hat{Y}^{(n)}_{ts}=\phi_{dec}(E_{s_{out}})
\end{equation}

\subsubsection{Time-Dynamic Koopman Prediction}
To capture the dynamic local variations, the time-dynamic Koopman predictor $KP_{td}$ leverages the weak stationarity of local time series segments. We divide the input sequence into multiple segments and apply the eDMD \cite{williams2015data} method to compute the time-dynamic Koopman operator $\mathcal{K}_{td}\in \mathbb{R}^{D \times D}$. Multi-step predictions are then performed iteratively in the Koopman embedding space.

We define the historical snapshot matrix as $E_{hist}=[\hat{e}_1,...,\hat{e}_{\frac{T}{L}-1}]$ and the future snapshot matrix as $E_{next}=[\hat{e}_2,...,\hat{e}_{\frac{T}{L}}]$. The time-dynamic Koopman operator $\mathcal{K}_{td}$ is obtained using the least squares method:

\begin{equation}
\mathcal{K}_{td}=E_{next}E_{hist}^{'}
\end{equation}

Here, $E_{hist}^{'}$ denotes the Moore-Penrose pseudoinverse of $E_{hist}$. $\mathcal{K}_{td} \in \mathbb{R}^{D \times D}$ is used to capture the dynamics within the current historical window. For the prediction length $H$, we start from the last observed embedding $\hat{e}_{\frac{T}{L}}$ and iteratively propagate forward to obtain $\frac{H}{L}$ predicted embeddings.

\begin{equation}
\tilde{e}_{\frac{T}{L}+t}=(\mathcal{K}_{td})^t\hat{e}_{\frac{T}{L}},t=1,2,...,H/L
\end{equation}

The predicted embeddings are then mapped back to the original data space via the decoder $\phi_{dec}$, generating the final prediction $\hat{Y}^{(n)}_{tv}$ for the time-dynamic component.

\begin{equation}
\tilde{x}_i=\phi_{dec}(\tilde{e}_i), \hat{Y}^{(n)}_{tv}=[\tilde{x}_{\frac{T}{L}+1},...,\tilde{x}_{\frac{T}{L}+\frac{H}{L}}]^T
\end{equation}

Meanwhile, we reconstruct the Koopman embedding of the input via $\mathcal{K}_{td}$, decoding it back as $\tilde{X}^{(n)}_{td}$.

\begin{equation}
[\tilde{e}_1,...\tilde{e}_{\frac{T}{L}}]=[\hat{e}_1,\mathcal{K}_{td}\hat{e}_1,...,\mathcal{K}_{td}\hat{e}_{\frac{T}{L}-1}]=[\hat{e}_1,\mathcal{K}_{td}E_{hist}]
\end{equation}

\begin{equation}
\tilde{X}_{td}^{(n)}=[\tilde{x}_1,...,\tilde{x}_{\frac{T}{L}}]^T=[\phi_{dec}(\tilde{e}_1),...,\phi_{dec}(\tilde{e}_{\frac{T}{L}})]^T
\end{equation}

\subsubsection{Optimization target}
We aim to minimize the MSE between the true and predicted values. We separately calculate the losses for the time-stable and time-dynamic components.

\begin{equation}
\begin{aligned}
L_{ts}&=\mathbb{E}_{Y^{(n)}}[||Y_{ts}^{(n)}-\hat{Y}_{ts}^{(n)}||^2], \\L_{td}&=\mathbb{E}_{Y^{(n)}}[||Y_{td}^{(n)}-\hat{Y}_{td}^{(n)}||^2]
\end{aligned}
\end{equation}

However, due to the notable pattern differences between the two components after multi-scale disentanglement, relying solely on independent modeling within each branch may neglect their potential interactions and collaborative relationships, leading to deviation of the optimization direction. To address this, we draw inspiration from ELBO optimization principle, aiming to jointly optimize the reconstruction errors of the sub-branches along with their residual correlations. We introduce a cross-term loss to explicitly model the synergy and complementary structure between the two components, thereby jointly enhancing the performance of both branches. This term ensures that the model accounts for the interdependence between the two components during prediction. The cross-term loss is defined as follows:

\begin{equation}
L_{cross}=2\mathbb{E}_{Y^{(n)}}[(Y_{ts}^{(n)}-\hat{Y}_{ts}^{(n)}) \cdot (Y_{td}^{(n)}-\hat{Y}_{td}^{(n)})]
\end{equation}

The total loss function combines of the above three losses:

\begin{equation}
L_{MSE}=L_{ts}+L_{td}+L_{cross}
\end{equation}

Expand the total loss function we obtained:

\begin{equation}
L_{MSE}=\mathbb{E}_{Y^{(n)}}||({Y}_{ts}^{(n)}+{Y}_{td}^{(n)})-(\hat{Y}_{ts}^{(n)}+\hat{Y}_{td}^{(n)})||^2
\end{equation}

We integrate all SDSTM blocks and obtain the total loss:

\begin{equation}
L_{MSE}=\mathbb{E}_{Y^{(n)}}||Y-\hat{Y}||^2
\end{equation}

$Y$ and $\hat{Y}$ represent the true and the predicted values. This loss function balances the impacts of both branches while considering their interaction. By minimizing it, SDSTM can construct the predicted sequence with higher accuracy.

\section{Experiments}
\subsection{Datasets}
We evaluate the rationality and effectiveness of SDSTM from multiple perspectives using a real-world vehicle emission dataset. It contains measurements of carbon monoxide (CO) concentrations across 1,298 valid road segments in the Second Ring Road area of Xi'an, located in Shaanxi Province, China. The spatial coverage is limited to the region bounded by longitude 108.92186–109.00934 and latitude 34.20495–34.27994. The road network structure is extracted from the OpenStreetMap database. The temporal range spans from October 1 to October 31, 2018, with 5-minute collected intervals. The missing rate of this dataset is only 0.11\%, indicating high reliability and completeness.

\subsection{Baselines}
For evaluation, we compared SDSTM with various advanced models for temporal prediction including LSTM \cite{graves2012long}, GRU \cite{cho2014learning}, Autoformer \cite{wu2021autoformer}, FEDformer \cite{zhou2022fedformer}, PatchTST \cite{nie2022time}, Koopa \cite{liu2023koopa}, TimeMixer \cite{wang2024timemixer}, TimeXer \cite{wang2024timexer}, and spatiotemporal prediction including STAEformer \cite{liu2023spatio}, STG-Mamba \cite{li2024stg}, PDG2Seq \cite{fan2025pdg2seq}. Check Appendix for details.

\subsection{Evaluation Metrics and Parameter Settings}

MSE and MAE are used to quantify the predictive accuracy. We split the datasets in the ratio of 7:2:1 respectively for training, testing and validation. Implementation is based on PyTorch 2.0 with an NVIDIA RTX4090 GPU. The model is trained using the Adam optimizer, starting with a learning rate of 0.001 that gradually decays to zero. The batch size is 32, and training runs for 10 epochs. To prevent overfitting, we use early stopping with a patience of 3. For the gated wavelet filter, we choose Daubechies 4 wavelet with a decomposition depth of 4. For each prediction window length H, we set the look-back window length $T = 2H$, except for TimeMixer and TimeXer, when $H=6$, we set $T=4H$.

\subsection{Performance Comparison}

\begin{table*}[!ht]
  \centering
  \setlength{\tabcolsep}{4.2pt}

  \renewcommand{\arraystretch}{0.8}
    \begin{tabular}{c|c|c|cccccc|c}
    \hline
    Models & Published & Metric & 6     & 12    & 24    & 48    & 96    & 144   & Improvement \bigstrut\\
    \hline
    \multirow{2}[2]{*}{LSTM} & \multirow{2}[2]{*}{2012} & MSE   & 0.41290  & 0.44322  & \textbf{0.47659} & 0.53953  & 0.66274  & 0.75710  & 17.20\% \bigstrut[t]\\
          &       & MAE   & 0.40428  & 0.42532  & \underline{0.44924}  & 0.49493  & 0.58093  & 0.64751  & 15.81\% \bigstrut[b]\\
    \hline
    \multirow{2}[2]{*}{GRU} & \multirow{2}[2]{*}{2014} & MSE   & 0.41871  & 0.44966  & \underline{0.49111}  & 0.54926  & 0.66940  & 0.76390  & 18.58\% \bigstrut[t]\\
          &       & MAE   & 0.41472  & 0.43681  & 0.45759  & 0.50251  & 0.58502  & 0.68910  & 17.85\% \bigstrut[b]\\
    \hline
    \multirow{2}[2]{*}{Autoformer} & \multirow{2}[2]{*}{NeurIPS 2021} & MSE   & 0.45350  & 0.53470  & 0.62680  & 0.72729  & 0.76724  & 0.59343  & 27.65\% \bigstrut[t]\\
          &       & MAE   & 0.43484  & 0.49860  & 0.55443  & 0.61929  & 0.64752  & 0.54570  & 24.27\% \bigstrut[b]\\
    \hline
    \multirow{2}[2]{*}{FEDformer} & \multirow{2}[2]{*}{ICML 2022} & MSE   & 0.44383  & 0.51157  & 0.62849  & 0.61923  & 0.60459  & 0.59573  & 22.16\% \bigstrut[t]\\
          &       & MAE   & 0.42258  & 0.47955  & 0.55701  & 0.56258  & 0.55363  & 0.54738  & 20.25\% \bigstrut[b]\\
    \hline
    \multirow{2}[2]{*}{PatchTST} & \multirow{2}[2]{*}{ICLR 2023} & MSE   & 0.34690  & 0.46492  & 0.63671  & 0.81114  & 0.69838  & 0.48693  & 19.60\% \bigstrut[t]\\
          &       & MAE   & 0.32844  & 0.40641  & 0.51423  & 0.63140  & 0.58865  & \underline{0.45142}  & 12.77\% \bigstrut[b]\\
    \hline
    \multirow{2}[2]{*}{Koopa} & \multirow{2}[2]{*}{NeurIPS 2023} & MSE   & 0.41429  & 0.45127  & 0.59819  & 0.66979  & 1.00920  & 1.34442  & 32.74\% \bigstrut[t]\\
          &       & MAE   & 0.38346  & 0.41829  & 0.50341  & 0.57092  & 0.65709  & 0.73470  & 21.66\% \bigstrut[b]\\
    \hline
    \multirow{2}[2]{*}{STAEformer} & \multirow{2}[2]{*}{CIKM 2023} & MSE   & 0.35687  & 0.47644  & 0.57750  & 0.67241  & 0.71085  & 0.68294  & 21.93\% \bigstrut[t]\\
          &       & MAE   & 0.34880  & 0.43944  & 0.50974  & 0.57923  & 0.61742  & 0.60188  & 18.41\% \bigstrut[b]\\
    \hline
    \multirow{2}[2]{*}{STG-Mamba} & \multirow{2}[2]{*}{2024} & MSE   & 0.49172  & 0.48916  & 0.51273  & \underline{0.52689}  & \underline{0.52987}  & 0.53152  & 13.95\% \bigstrut[t]\\
          &       & MAE   & 0.44924  & 0.44956  & 0.46657  & \underline{0.47729}  & \underline{0.46981}  & 0.47675  & 10.53\% \bigstrut[b]\\
    \hline
    \multirow{2}[2]{*}{TimeMixer} & \multirow{2}[2]{*}{ICLR 2024} & MSE   & 0.32443  & 0.45210  & 0.59265  & 0.65495  & 0.54228  & \textbf{0.47419} & 11.05\% \bigstrut[t]\\
          &       & MAE   & 0.31707  & 0.40047  & 0.49212  & 0.55454  & 0.49993  & \textbf{0.44625} & 7.12\% \bigstrut[b]\\
    \hline
    \multirow{2}[2]{*}{TimeXer} & \multirow{2}[2]{*}{NeurIPS 2024} & MSE   & 0.34163  & 0.46247  & 0.57773  & 0.59577  & 0.55954  & 0.48511  & 11.44\% \bigstrut[t]\\
          &       & MAE   & 0.33528  & 0.41145  & 0.48532  & 0.52581  & 0.51260  & 0.45647  & 8.19\% \bigstrut[b]\\
    \hline
    \multirow{2}[2]{*}{PDG2Seq} & \multicolumn{1}{c|}{\multirow{2}[2]{*}{Neural Networks 2025}} & MSE   & \textbf{0.28731} & \textbf{0.34708} & 0.54004  & 0.66756  & 0.77252  & 0.88928  & 15.69\% \bigstrut[t]\\
          &       & MAE   & \textbf{0.28703} & \textbf{0.33528} & 0.46073  & 0.55212  & 0.63087  & 0.69627  & 10.68\% \bigstrut[b]\\
    \hline
    \multirow{2}[2]{*}{SDSTM} & \multirow{2}[2]{*}{-} & MSE   & \underline{0.31218}  & \underline{0.40410}  & 0.49219  & \textbf{0.48732} & \textbf{0.48192} & \underline{0.48256}  & - \bigstrut[t]\\
          &       & MAE   & \underline{0.30285}  & \underline{0.37399}  & \textbf{0.44089} & \textbf{0.46410} & \textbf{0.46460} & 0.45587  & - \bigstrut[b]\\
    \hline
    \end{tabular}%
    \caption{Main results}
  \label{tab:main results}%
\end{table*}%

We extensively compare SDSTM with current state-of-the-art temporal forecasting models and spatiotemporal forecasting models. We repeated each experiment three times using different random seeds and average the results as reported in Table \ref{tab:main results}. We highlight the best results in bold and the second-best results in underline. The last column is the average percentage lift of SDSTM relative to the method of each row, we calculate the improvement of the corresponding metrics separately for each prediction length setting and compute the mean value. The inferences are as follows:

\begin{itemize}
    \item At short prediction length settings, the spatiotemporal method PDG2Seq presents a better performance than those pure temporal methods, which seems to be natural: combining spatial information can capture the correlation between nodes in a more direct way, complementing the lack of information in short time series itself.
    \item For long prediction step setting, pure temporal prediction methods like TimeMixer gradually show their advantages, due to the fact that they focus more on mining the long-term dependencies in time series. As prediction step increases, the high-frequency spatiotemporal dynamics gradually evolve into random signals, and the dynamics modeling of the spatiotemporal methods lacks predictability, thus exhibits deterioration in accuracy.
    \item As prediction step increases, accuracy of each method drops to different degrees, only STG-Mamba and SDSTM present considerable stability across all prediction horizons, exhibits stronger resistance to error accumulation. Meanwhile, SDSTM achieves better accuracy.
    \item In general terms, SDSTM not only achieves optimal or suboptimal results for various prediction length settings, but also solves the performance decay problem commonly found in long term, and exhibits superior performance in medium and long term tasks. This suggests that SDSTM is able to balance learning the global evolutionary trends and capturing the local changing features, demonstrating its significant spatiotemporal dynamic modeling capability in long term prediction tasks.
\end{itemize}

\subsection{Ablation Study}
To thoroughly validate the contribution of all constituent modules, we conducted an ablation study of SDSTM:

\begin{itemize}
    \item \textbf{w/ Wavelet}: This variant removes the spatial module, meaning it only uses information of time dimension.
    \item \textbf{w/ Fourier}: This variant has the same structure as \textit{w/ Wavelet}, but replaced gated wavelet filter with Fourier Filter adopted from Koopa \cite{liu2023koopa}.
    \item \textbf{w/o SD}:This variant removes Spatial Dynamic module.
    \item \textbf{w/o SS}: This variant removes Spatial Stable module .
\end{itemize}

\begin{table}[!b]
  \centering

  \setlength{\tabcolsep}{1mm}
  \resizebox{0.49\textwidth}{!}{
    \begin{tabular}{ccccccccccccc}
    \hline
    \multirow{2}[4]{*}{Pred Len} & \multicolumn{2}{c}{6} & \multicolumn{2}{c}{12} & \multicolumn{2}{c}{24} & \multicolumn{2}{c}{48} & \multicolumn{2}{c}{96} & \multicolumn{2}{c}{144} \bigstrut\\
\cline{2-13}          & MSE   & MAE   & MSE   & MAE   & MSE   & MAE   & MSE   & MAE   & MSE   & MAE   & MSE   & MAE \bigstrut\\
    \hline
    w/ Wavelet & 0.33  & 0.32  & 0.40  & 0.38  & 0.49  & 0.44  & 0.48  & 0.46  & 0.59  & 0.54  & 0.67  & 0.56  \bigstrut[t]\\
    w/ Fourier & 0.41  & 0.38  & 0.45  & 0.42  & 0.60  & 0.50  & 0.67  & 0.57  & 1.01  & 0.66  & 1.34  & 0.73  \bigstrut[b]\\
    \hline
    w/o SD & 0.32  & 0.31  & 0.42  & 0.39  & 0.49  & 0.44  & 0.49  & 0.47  & 0.48  & 0.47  & 0.49  & 0.46  \bigstrut[t]\\
    w/o SS & 0.32  & 0.30  & 0.40  & 0.38  & 0.50  & 0.45  & 0.50  & 0.47  & 0.50  & 0.48  & 0.47  & 0.46  \bigstrut[b]\\
    \hline
    SDSTM & 0.31  & 0.30  & 0.40  & 0.37  & 0.49  & 0.44  & 0.49  & 0.46  & 0.48  & 0.46  & 0.48  & 0.46  \bigstrut\\
    \hline
    \end{tabular}}%
    \caption{Ablation Study}
  \label{tab:table2}%
\end{table}%

The following conclusions can be inferred from Table \ref{tab:table2}: (1) The original SDSTM consistently exhibits excellent performance across missions, demonstrating the effectiveness of its complete configuration. (2) \textit{w/ Wavelet} consistently outperforms \textit{w/ Fourier}, which demonstrates that our designed wavelet filter has better results for decomposing the time-stable components and time-dynamic components. (3) The overall performance for \textit{w/ Wavelet} are not as good as the other variants that employ spatial module, claiming the key role of spatial information. (4) The results of \textit{w/o SD} and \textit{w/o SS} show alternating dominance at different prediction step settings, which indicates that both dynamic graph module and static graph module are indispensable.

\subsection{Model Visualization Study}
\subsubsection{Temporal Dimension}
\begin{figure}[!t]
\centering
\includegraphics[width=3.2in]{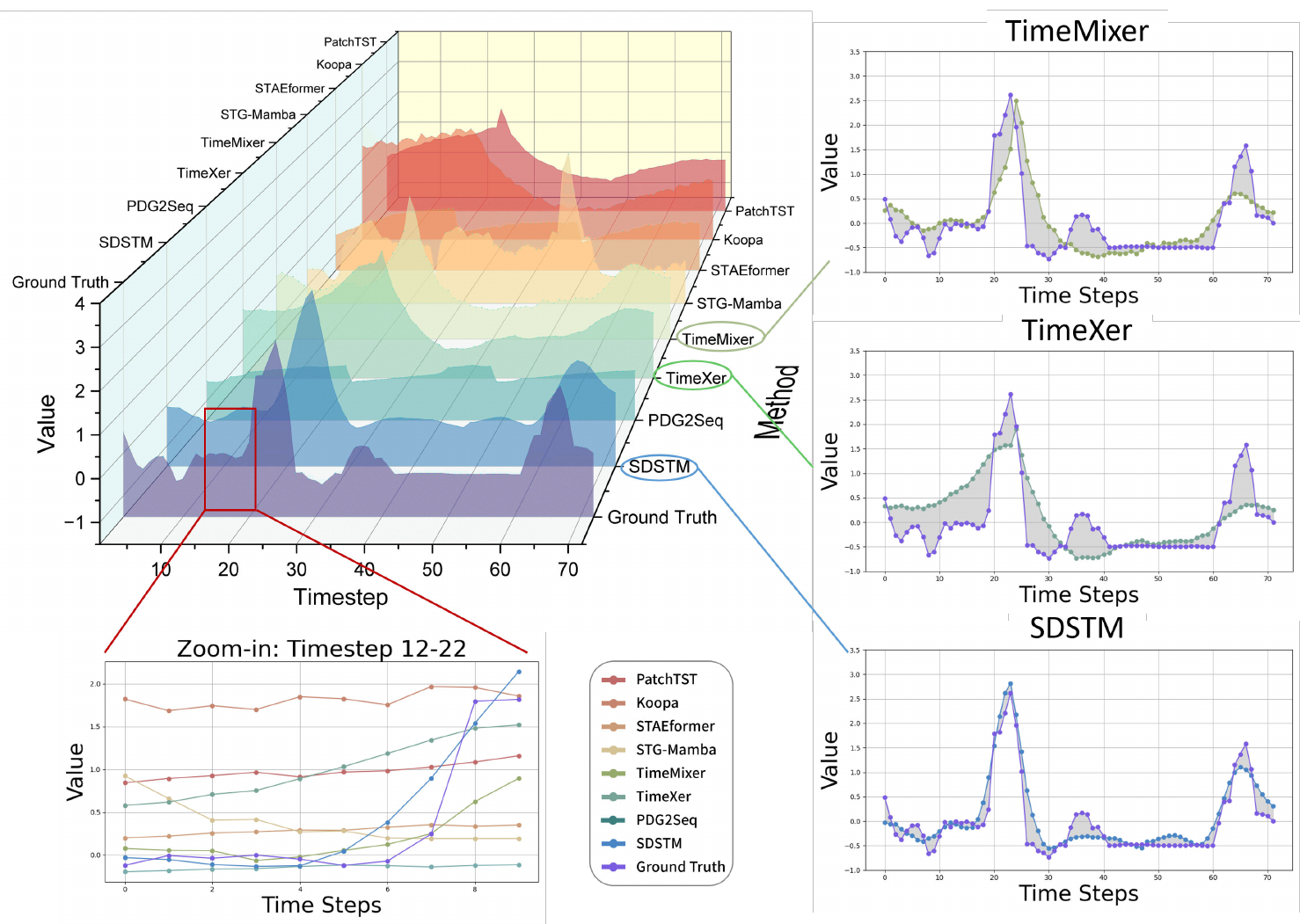}
\caption{Visualization of the true values and the predicted values of CO emissions of different methods (H=96).}
\label{fig:waterfall}
\end{figure}

We visualize predicted and true values for a typical road segment over October 23, 2018 in Fig. \ref{fig:waterfall}, with 20-minute averaged, normalized data.

By and large, SDSTM accurately predicts peak timings and better captures morning and evening emission peaks.
To make it easier to observe, in the right column, we demonstrate the top-3 models by prediction performance for the prediction step setting of this experiment (ranking from Table \ref{tab:main results}.), plotting the predicted value of the method versus the true value in each subfigure, connecting the data discrepancy between each time step as a shaded image to more intuitively show the average absolute error of model. As we can see, SDSTM shows the smallest, evenly distributed shaded area, with predicted trends closely matching the true values.

To show details, we zoom in on the 12-22 time step, in the lower left corner, around 4:00–7:30 a.m., marked by the red box in main figure. Compared to other methods, SDSTM not only closely follow the sequence when the trend is relatively flat, but also manages to react with timely forecasts when the sequence shows a clear upward trend. This proves that in the time dimension, our method not only has strong global trend modeling ability, but also can capture local abrupt changes.

\subsubsection{Spatial Dimension}

\begin{figure}[!t]
\centering
\includegraphics[width=0.8\linewidth]{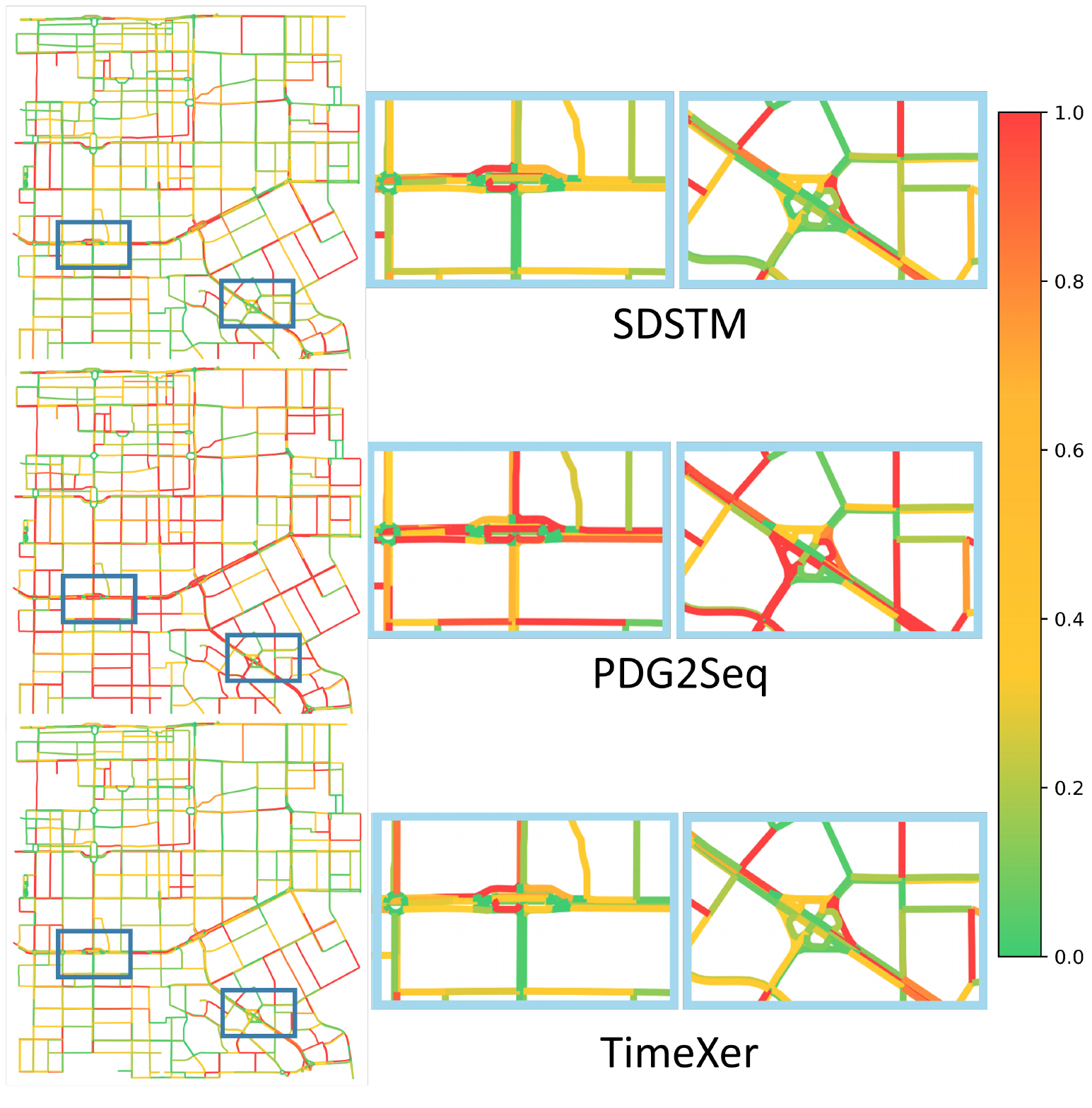}
\caption{Visualization of the mapping of prediction errors of each road segment for long prediction length (H=96) .}
\label{fig:Load_96}
\end{figure}

To illustrate that SDSTM can model spatial features well, for long-term prediction, we map the morning peak hour (8:00–9:00) prediction error onto the road network (Fig. \ref{fig:Load_96}) and zoom in on the transportation hub. We calculate MSE on normalized data within the certain time period. Color of road reflects its prediction accuracy. We compare SDSTM with one each of typical spatiotemporal methods (PDG2eq) and temporal methods (TimeXer).

For long prediction steps setting, PDG2Seq shows a large red area in Fig. \ref{fig:Load_96}, indicating that although it can model spatial features to a certain extent, it lacks the ability to learn dynamic spatial evolution, as prediction steps increase, error aggregation occurs in some local region. TimeXer shows better performance, but the prediction effect for global and the focused transportation hub area is not as good as SDSTM, which means that TimeXer as a time series method lacks the ability to learn complex spatial structure. In summary, SDSTM exhibits strong spatiotemporal dynamic modeling capability in long-term prediction task.


\subsection{Model Efficiency Study}
We evaluate model efficiency through three dimensions including prediction performance, GPU memory usage and training speed as shown in Fig. \ref{fig:bubbles}. Horizontal axis represents the maximum GPU memory usage while training, and vertical axis represents the prediction accuracy derived from Table. \ref{tab:main results}. Bubbles represent the time consumed for each iteration, larger and darker bubbles represents more consumed time. From Fig. \ref{fig:bubbles}, we can observe that SDSTM demonstrates advanced overall performance in balancing prediction accuracy and computational resource consumption.


\subsubsection{Time Distangle Analysis}
\begin{figure}[!t]
\centering
\subfigure[]{\includegraphics[width=0.24\textwidth]{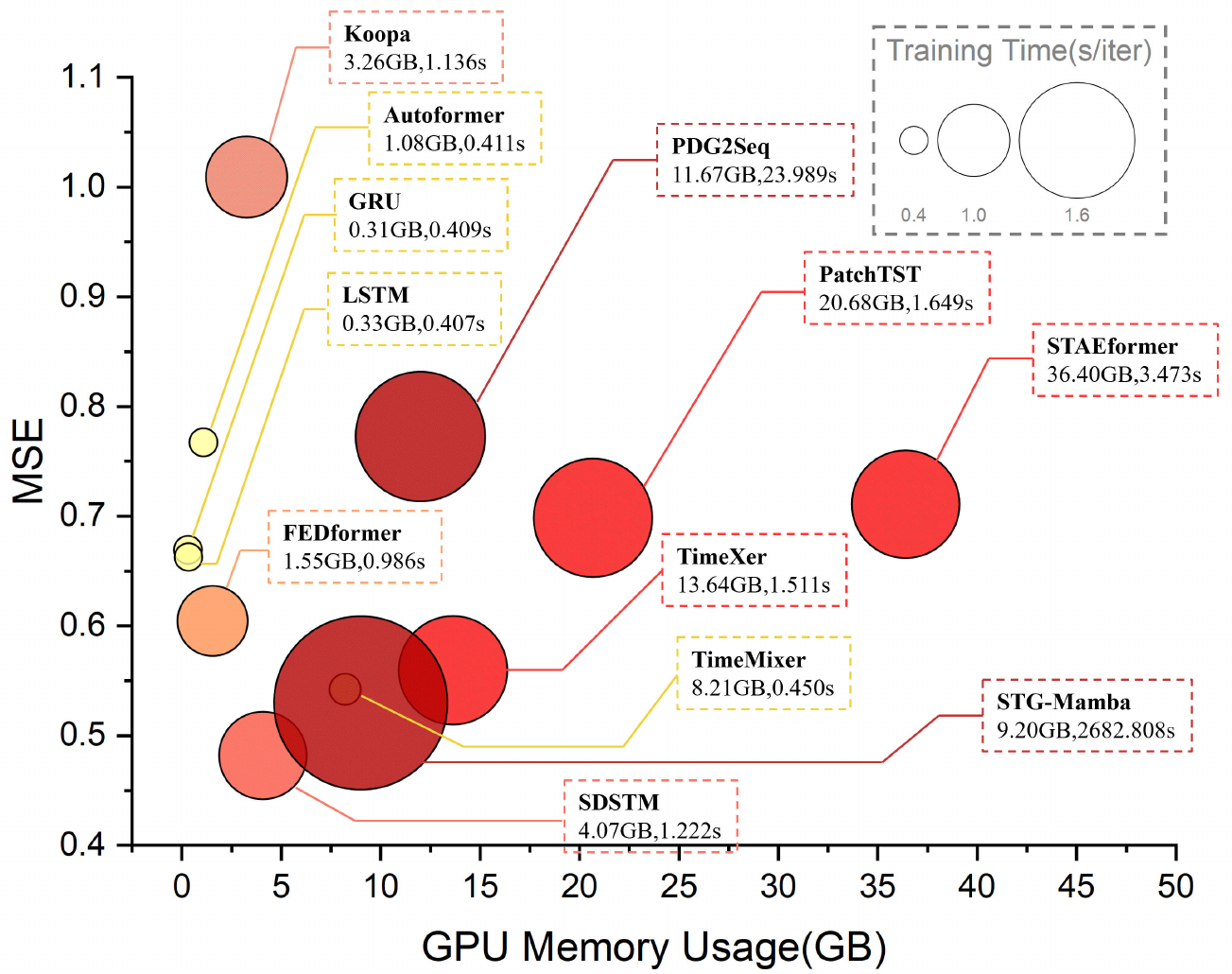}
\label{fig:bubbles}}
\hspace{-0.15in}
\subfigure[]{\includegraphics[width=0.232\textwidth]{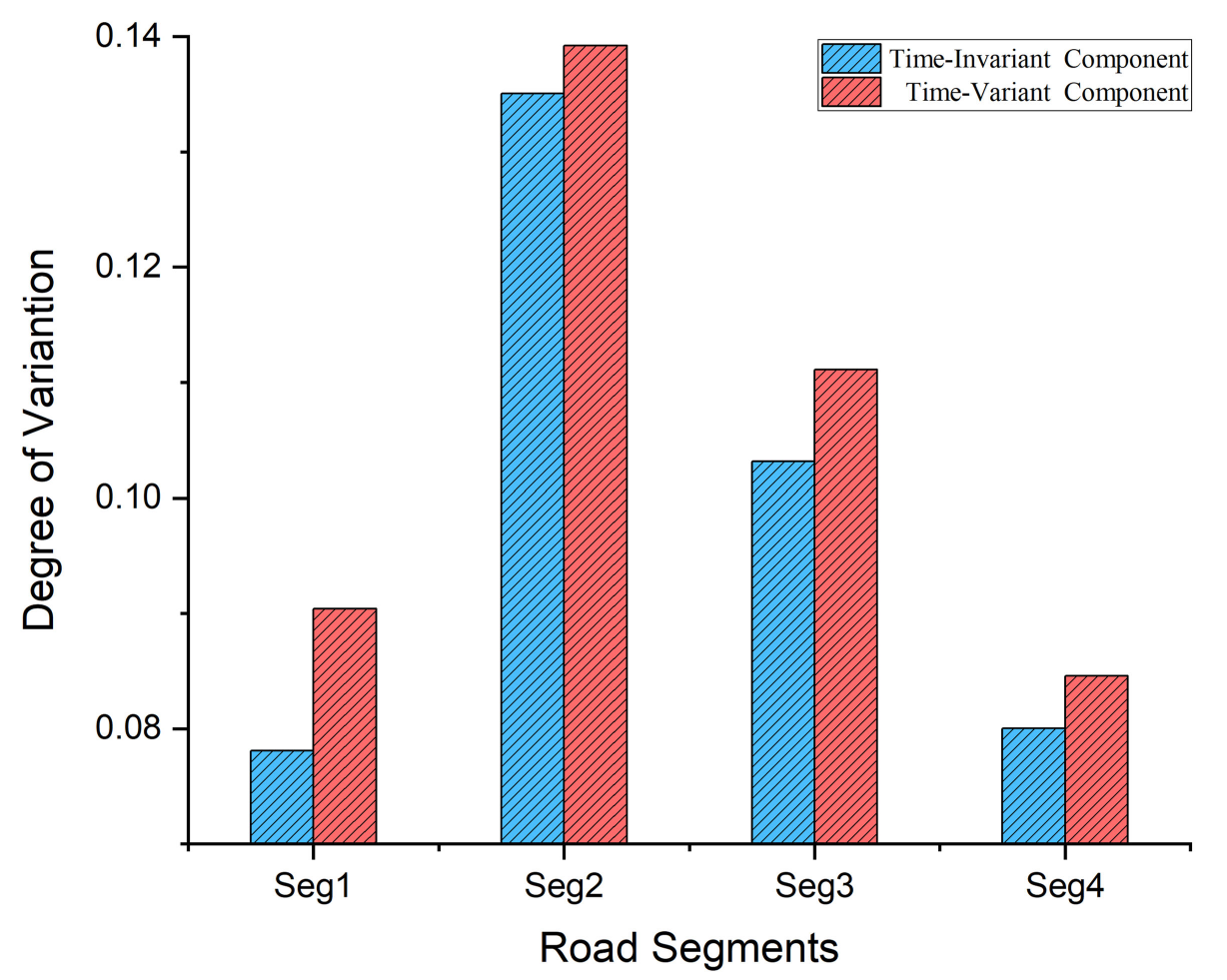}
\label{fig:timedistangle}}
\caption{(a) Model efficiency comparison (H=96) . (b) Comparison of the mean standard deviation of linear fits to different subsets of cycles of random road segments.}
\label{fig:ModelAnalysis}
\end{figure}

To presents the effectiveness of gated wavelet filter, we adopt the degree of variation metric proposed in \cite{liu2023koopa}. This metric applies the filter to 20 periodic subsets of one dataset, and separately performing linear regression on the two decomposed components. The standard deviation of the regression results is used to quantify the degree of fluctuation over time. We randomly select four road segments data for testing. Results shows in Fig. \ref{fig:timedistangle} that the time-dynamic components exhibit stronger fluctuation than the time-stable components, indicating that our module successfully separates them.

\section{Conclusion}
In this study, we propose a scale-decoupled spatiotemporal modeling framework to address the challenges of long-term prediction caused by the multiscale entanglement inherent in traffic emission. SDSTM consists of a dual-stream feature decomposition strategy and an ELBO fusion mechanism. The former integrates Koopman theory with gated wavelet to achieve linear embedding and multiscale decoupling of spatiotemporal dynamical systems, effectively delineating the predictability boundary and enhancing the model’s ability to analyze complex spatiotemporal evolution. The latter introduces a dual-stream independence constraint based on a cross-term loss, dynamically refining the predictions by suppressing mutual interference and improving the accuracy of long-term traffic emission forecasting. Our model demonstrates competitive performance in experiments and effectively alleviates the degradation of spatiotemporal model performance in long-term prediction tasks.

\bibliography{main}

\end{document}